\documentclass[letterpaper, 10 pt, conference]{ieeeconf}  

\IEEEoverridecommandlockouts                              

\usepackage{multicol}
\usepackage[bookmarks=true]{hyperref}
\usepackage[ruled]{algorithm2e}
\usepackage{algpseudocode}
\usepackage{graphics}
\usepackage{amsmath}

\usepackage{amssymb}
\usepackage{epsfig} 
\usepackage{booktabs}
\usepackage{float}
\usepackage{url}
\usepackage{gensymb}
\usepackage{hyperref}
\hypersetup{
    colorlinks=true,
    urlcolor=red,
    }
\usepackage{makecell}
\usepackage[colorinlistoftodos]{todonotes}
\newcount\Comments 
\usepackage{array,multirow,graphicx}
\usepackage{color}
\usepackage{wrapfig}
\definecolor{darkgreen}{rgb}{0,0.5,0}
\definecolor{purple}{rgb}{1,0,1}
\newcommand{\kibitz}[2]{\ifnum\Comments=1\textcolor{#1}{#2}\fi}

\title{\LARGE \bf
Dynamic Grasping with a Learned Meta-Controller
}

\author{Yinsen Jia$^{*, 1}$, Jingxi Xu$^{*, 1}$, Dinesh Jayaraman$^2$, Shuran Song$^1$ \\ $^*$ Equal Contribution \quad\quad $^1$ Columbia University \quad\quad  $^2$ University of Pennsylvania
\\ \href{https://yjia.net/meta}{https://yjia.net/meta}}

\begin{document}

\maketitle
\thispagestyle{empty}
\pagestyle{empty}

\begin{abstract}
Grasping moving objects is a challenging task that requires multiple submodules such as object pose predictor, arm motion planner, etc. Each submodule operates under its own set of meta-parameters. For example, how far the pose predictor should look into the future (i.e., \textit{look-ahead time}) and the maximum amount of time the motion planner can spend planning a motion (i.e., \textit{time budget}). Many previous works assign fixed values to these parameters; however, at different moments \textit{within} a single episode of dynamic grasping, the optimal values should vary depending on the current scene. In this work, we propose a dynamic grasping pipeline with a meta-controller that controls the look-ahead time and time budget dynamically. We learn the meta-controller through reinforcement learning with a sparse reward. Our experiments show the meta-controller improves the grasping success rate (up to 28\% in the most cluttered environment) and reduces grasping time, compared to the strongest baseline. Our meta-controller learns to reason about the reachable workspace and maintain the predicted pose within the reachable region. In addition, it assigns a small but sufficient time budget for the motion planner. Our method can handle different objects, trajectories, and obstacles. Despite being trained only with 3-6 random cuboidal obstacles, our meta-controller generalizes well to 7-9 obstacles and more realistic out-of-domain household setups with unseen obstacle shapes.
\end{abstract}
\section{Introduction}
\label{sec:intro}

Grasping moving targets without prior knowledge of their motion remains a challenging task, requiring many trade-offs to be made in real-time. For example, the robot needs accurate sensing and extended planning given sufficient time, but coarse, fast sensing and replanning under time pressure.  
We decompose the entire task into several submodules. A vision system identifies the target object to grasp and estimates its pose. An object pose predictor predicts an object's future poses. A grasp planner plans a grasp for the object at this predicted pose. Finally, an arm motion planner plans a collision-free motion to reach that grasp. The above processes make up one iteration of dynamic grasping, and the loop repeats until the object is close enough to the end-effector to be grasped. 


Each of these submodules operates under its own set of meta-parameters, and we propose to dynamically control the two most representative meta-parameters: \textit{look-ahead time} for the object pose predictor and \textit{time budget} for the arm motion planner. 
The interplay among them brings a lot of interesting trade-offs into this system and their assignments impact the performance dramatically. Many previous works~\cite{akinola2021dynamic, ye2018velocity, kappler2018real, marturi2019dynamic, allen1993automated} assign a fixed value to these parameters; however, during dynamic grasping, their optimal values should differ at each iteration, depending on the current scene/environment.

\begin{figure}
    \centering
    \includegraphics[width=\linewidth]{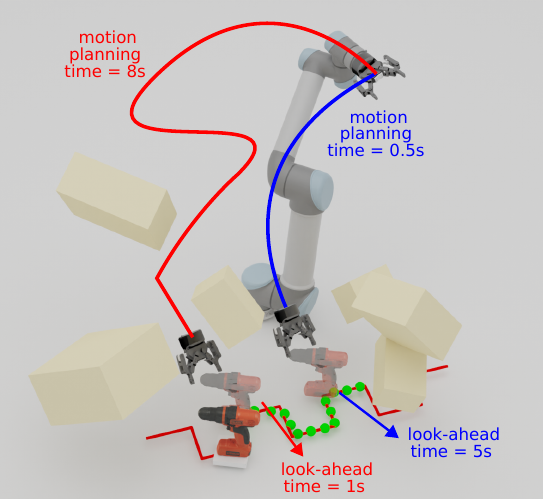}
    \vspace{-5mm}
    \caption{\footnotesize\textbf{Meta-parameters in dynamic grasping.} The green dots indicate the predicted trajectory for the whole predictable range (8s). The red periodic rectangular line indicates the object's trajectory. The semi-transparent objects indicate the predicted poses at different look-ahead times. If the robot uses a large look-ahead time (5s), the predicted pose will fall into a highly reachable space where the motion planner can plan a motion fast (0.5s). If the robot uses a small look-ahead time (1s), the predicted pose will locate in a highly cluttered area where a collision-free path takes longer to plan (8s). 
    In this example, our meta-controller uses a large look-ahead time and a small time budget. The robot can immediately plan a collision-free motion and move.}
    \label{fig:teaser}
    \vspace{-6mm}
\end{figure}

Look-ahead time controls how far the pose predictor looks into the future. The object pose predictor can predict a pose that is 3 seconds away, 6 seconds away, etc., which equals the value of the look-ahead time. Prediction is critical in dynamic grasping because, without prediction, the robot arm can never catch up with the target object. Many robotic dynamic tasks rely on object pose predictors, but very few study the trade-off of the look-ahead time. If we predict a pose that is very far from the current object (large look-ahead time), we have more time reserved for motion and grasp planning, but worse prediction accuracy. In addition, the look-ahead time controls the predicted object pose and, consequently, affects the success of the motion planning. For example (as shown in Fig.~\ref{fig:teaser}), if the object has moved into a cluster of obstacles, the pose predictor should use a large look-ahead time to generate a future pose outside these obstacles for easier motion planning.

The time budget for a motion planner is the maximum amount of time allowed for planning a motion. If no successful path is planned within the time budget, the motion planning is terminated with a failure. Otherwise, a solution is returned as soon as a successful path is planned. It brings the trade-off between planning success and delay. The larger the time budget, the more likely a successful path can be found. But if a large time budget is used up, the target object might have already moved too far away.


In summary, our contributions are: (1) To the best of our knowledge, we are the first to propose a dynamic grasping pipeline with a learned meta-controller. We identify and dynamically control two critical meta-parameters: the look-ahead time and the time budget. 
(2) We propose to train the meta-controller through reinforcement learning with a sparse reward. Our meta-controller significantly improves the success rate (up to 28\% in the most cluttered environment) and reduces the grasping time. Despite being trained only with 3-6 random cuboidal obstacles, our meta-controller generalizes well to more obstacles and more realistic household setups with unseen obstacle shapes. (3) We propose a simulation benchmark that covers a variety of trajectories and obstacles to evaluate the performance of dynamic grasping algorithms. 


\section{Related Work}

\paragraph{Dynamic Grasping} There are many attempts at the problem of dynamic grasping, but many of them make some assumptions, such as a known object moving trajectory~\cite{allen1993automated, hu2021learning}, top-down grasps only~\cite{ye2018velocity, wong2022moving, app112110270}, slow and small object motion~\cite{burgess2022eyes,burgess2022dgbench}, human-signaled grasp execution~\cite{kappler2018real}, access to highly effective reactive controllers or motion planners~\cite{fishman2023motion, bhardwaj2022storm, qureshi2020motion, cheng2020rmp, van2022geometric}, etc. Many works do not consider obstacles or only have a few obstacles with a large open workspace for the robot arm~\cite{marturi2019dynamic, akinola2021dynamic, wu2022grasparl, jang2022vision, cowley2013perception}. In these works, the meta-parameters of each submodule are often assigned heuristically or experimentally through grid search. They all require committing to the same meta-parameters for a full object-grasping episode or even holding the same meta-parameters for all episodes. Instead, the focus of this work is on modulating these parameters at each moment \textit{within} each episode. 

\paragraph{Delay-accuracy Trade-off}
The trade-off between computation time (delay) and solution qualities (accuracy) is critical in autonomous systems for real-time planning and control problems. Early work has concentrated on systems with perfect sensors and effectors and with unlimited computational power, which is impractical in most real-time applications. Anytime algorithms~\cite{Horvitz1987ReasoningAB, dean1988analysis, zilberstein1996using}, whose accuracy improves gradually as computation time increases, can be interrupted at any time to provide a valid solution that the rest of the system can act on. The flexibility offered by anytime algorithms allows accurate sensing and extended planning when time is available, and coarse, fast sensing and planning under time pressure. This flexibility makes real-time robotic applications possible and is widely used in a wide range of topics, including motion planning~\cite{karaman2011anytime}, object detection~\cite{karayev2014anytime}, probabilistic inference~\cite{ramos2005anytime}, belief space planning~\cite{pineau2003point, spaan2005perseus}, etc. However, anytime algorithms bring a new trade-off between solution quality and computation time, and to optimize this trade-off, autonomous systems have to solve a meta-level control problem; namely, they must decide when to interrupt the anytime algorithm and act on the current solution. 

There are two main approaches to solving the accuracy-delay trade-off. \textit{Fixed allocation} runs the anytime algorithm until a stopping point which is determined prior to runtime~\cite{horvitz2013reasoning, boddy1994deliberation}. These approaches are fragile when there is large uncertainty presented in the anytime algorithm. \textit{Monitoring and control}~\cite{horvitz1991computation, zilberstein1995approximate, hansen2001monitoring, lin2015metareasoning, pant2020anytime, zilberstein1993anytime, pant2015co} relies on a performance profile that monitors the performance of the algorithm at runtime and adjusts the stopping point periodically. This performance profile must be compiled offline before the activation of meta-level control. 
Relying on such a profile imposes many assumptions that are hard to guarantee, such as the same settings of the anytime algorithm across all problem instances and known/fixed distribution of problem instances.

\cite{falanga2019fast} studies perception latency specifically, and \cite{Sripathy2021-gb} uses computationally inefficient models selectively. \cite{graves2016adaptive} studies the adaptive computation time for neural networks. \cite{svegliato2020model} uses reinforcement learning to achieve meta-level control. However, most of the above works study this trade-off on a mobile robot for navigation tasks or they focus on one specific meta-parameter, but we demonstrate our meta-controllers on a more complex dynamic grasping task, with multiple submodules and meta-parameters. Our meta-controller assigns two meta-parameters simultaneously, whose interplay presents a more challenging problem than a single meta-parameter. 

\section{Method}

\begin{figure*}
    \centering
    \includegraphics[width=0.98\linewidth]{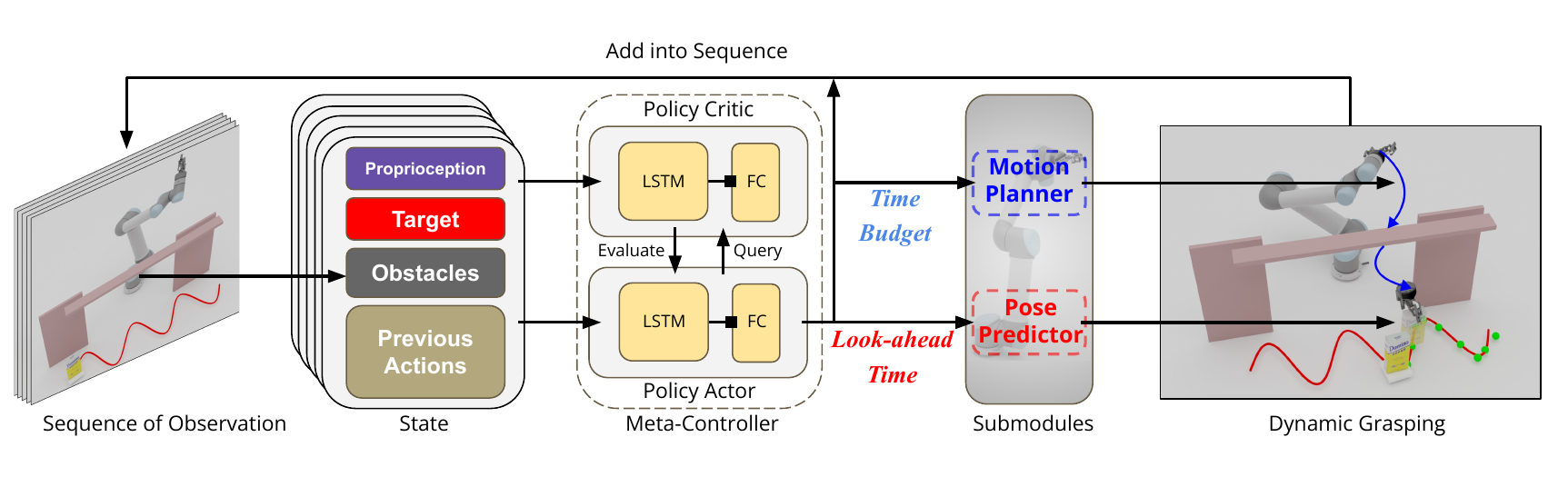}
    \vspace{-5mm}
    \caption{\footnotesize\textbf{Dynamic grasping with a learned meta-controller.} 
    We stack a sequence of scene information from the past 5 iterations as the state input to our meta-controller. 
    Our meta-controller generates for the current iteration a look-ahead time $T_L$ for the object pose predictor, and a time budget $T_T$ for the motion planner. 
    We model the meta-controller as a PPO agent trained with RL and a sparse reward. 
    }
    \label{fig:pipeline}
    \vspace{-6mm}
\end{figure*}



At the core of our method is a dynamic grasping pipeline with multiple submodules and a learned meta-controller. The goal of the meta-controller is to dynamically assign the look-ahead time and time budget as discussed in Sec.~\ref{sec:intro}. Assigning the correct values is difficult, and even humans can not provide optimal and straightforward intuitions. The assignment of these meta-parameters has a delayed effect, and the interplay between them is hard to model.


\subsection{Dynamic Grasping Pipeline}
Our dynamic grasping pipeline consists of an object pose predictor, a grasp planner, and an arm motion generator. The goal of this work is not to develop standalone components that outperform previous works in their specific jobs. Instead, we want to improve the overall performance of dynamic grasping by dynamically controlling the meta-parameters to achieve a more efficient interplay among them.

For each iteration of dynamic grasping, a vision system identifies the target object and estimates its pose. 
In the simulation, we apply a noise $\sim \mathcal{N}(0, 10\text{mm})$ on the object poses to simulate errors of the real-world pose estimator. An object pose predictor takes in a sequence of past estimated poses and generates a predicted pose according to the look-ahead time. 
After the predicted pose is determined, the grasp planner plans a 6DOF (degrees of freedom) grasp for the object at this predicted pose. The arm motion planner then plans a collision-free motion to reach that grasp, within the specified time budget.
The robot arm starts moving and the loop continues to the next iteration until the end-effector is close enough to the target for grasp execution. The detailed implementation for each component is as follows.

\textbf{Object Pose Predictor}. 
The input to our pose predictor is a sequence of past poses and velocities, sampled from the past 12 seconds at 16Hz. It also takes a look-ahead time within the predictable range of 8 seconds. The output is the predicted pose at the look-ahead time. All poses are relative to their previous pose, such that the pose predictor can easily generalize to trajectories at different locations. We collect a training dataset by sampling various trajectories in simulation. 
Details on the network architecture can be found in the appendix.


\textbf{Grasp Planner}. 
For each object, we pre-generate 5 grasp poses using GraspIt!~\cite{miller2004graspit}. We ensure that these five grasps are diverse from each other to cover as many orientations as possible. During grasp planning, we first check if the grasp pose from the previous iteration still has a valid IK. If not, we rank the other 4 grasps according to their $l_2$-norm distances to the current robot end-effector pose. From the closest to the furthest, we choose the first grasp with an IK solution. 

\textbf{Arm Motion Planner}. We use BiRRT. Given a time budget, our motion planner returns the solution as soon as a valid collision-free motion is found; or it returns failure after the time budget is used up. The planned motion is then retimed to match the real-robot speed. 

\subsection{Learning a Meta-controller}
In our method, the meta-controller is modeled as an RL agent. Without any prior knowledge, our meta-controller learns to control the meta-parameters with a sparse reward.

\subsubsection{State and Action Space}
It is critical for our meta-controller to know enough about the current scene.
Through a variety of ablation experiments, we find that the below information is important to learn a meta-controller, as shown in Fig.~\ref{fig:pipeline}. (1) Proprioception: the robot arm joint angles $\in \mathbb{R}^n$, where $n=6$ is the number of degrees of freedom, and the end-effector pose (position + quaternion) $\in \mathbb{R}^7$. (2) Target: the target object moving speed $[v_x, v_y, v_z] \in \mathbb{R}^3$ computed using past estimated poses, the distance from the object to our arm end-effector, the target pose $\in \mathbb{R}^7$, and target bounding box dimension $[d_x, d_y, d_z] \in \mathbb{R}^3$. This also includes the predicted target pose and the distance between the predicted target pose to the arm end-effector. (3) Obstacles: similar to the target object, we model the 5 closest obstacles to the end-effector with a 10-dimension vector (position + quaternion + bounding box dimension). 
We apply a noise $\sim \mathcal{N}(0, 10\text{mm})$ to the ground-truth obstacle poses and bounding box dimensions. (4) Previous Actions: this contains all the information about the meta-parameters (look-ahead time, time budget), motion planning (motion planning success, motion planning time) and grasp planning (grasp planning success, planned grasp pose, grasp planning time, IK configuration of planned grasp pose, distance from the arm configuration to the IK configuration of the planned grasp) in the previous iteration. 

Our meta-controller takes in a stacked history of such information from the past 5 iterations and produces the meta-parameters for the current iteration. The meta-controller outputs two continuous actions, look-ahead time $T_L \in [0s, 8s]$ and motion planning time budget $T_T \in [0s, 4s]$.

\begin{figure}[b]
    \centering
    \vspace{-5mm}
    \includegraphics[width=\linewidth]{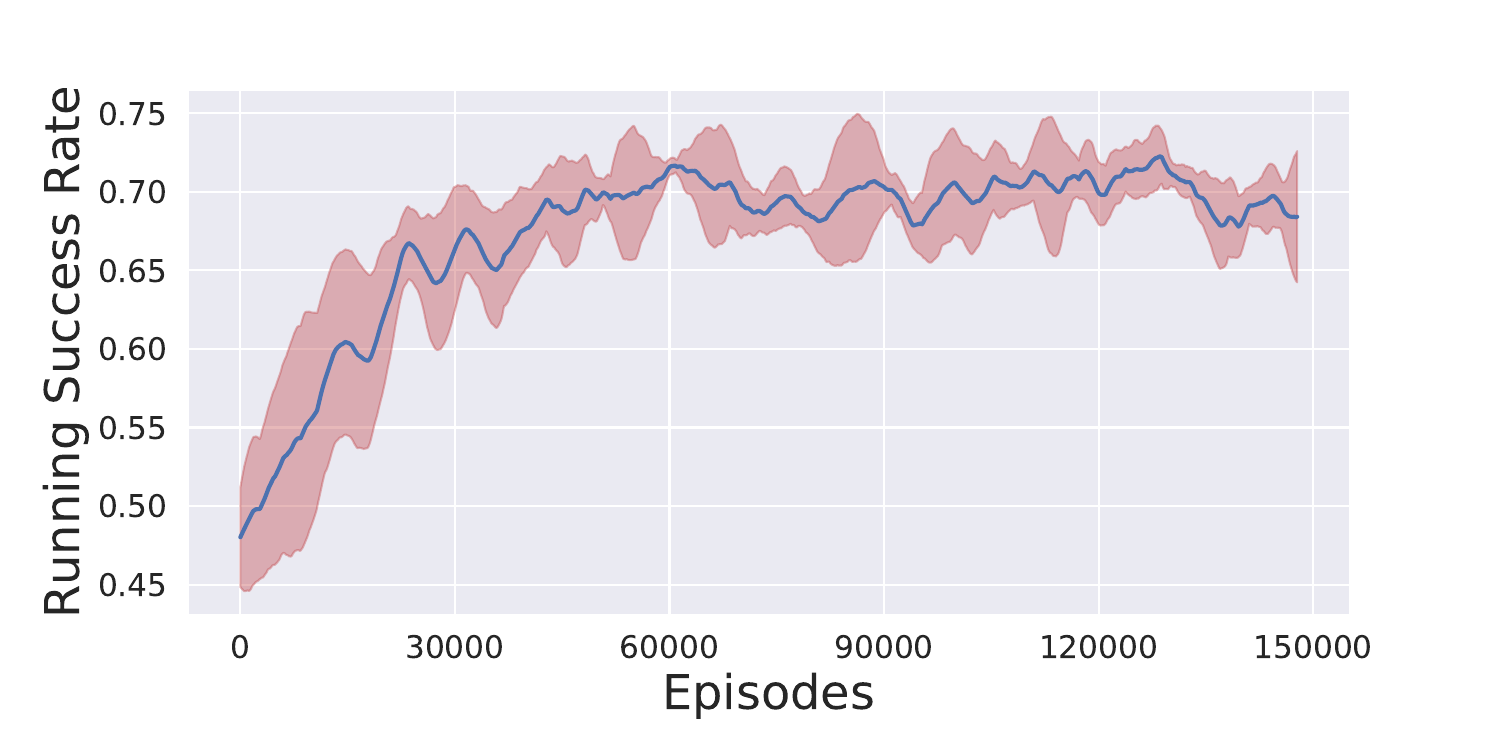}
    \vspace{-5mm}
    \caption{\footnotesize\textbf{Meta-controller training plot.} Our meta-controller is trained up to 150,000 episodes (around 24 hours) with 5 random seeds. We visualize the mean of the running success rate over the past 1000 episodes and 1 standard deviation is shaded.}
    \label{fig:trainin_plot}
\end{figure}

\begin{figure*}
\begin{center}
    \centering
    \includegraphics[width=\linewidth]{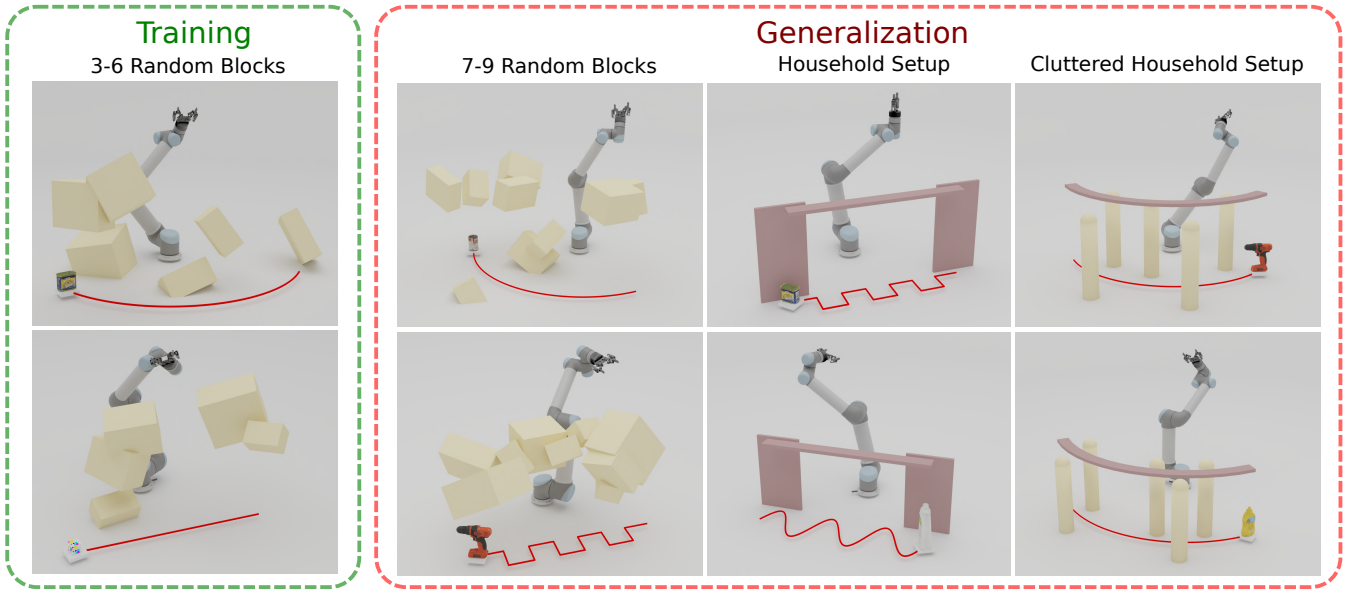}
    \vspace{-5mm}
    \caption{\footnotesize\textbf{Experimental setups.} We show two randomly selected examples for each of our 4 setups. 
    In each example, the target object is on the conveyor at the start of the trajectory. The robot arm is in its random initial configuration. 
    }
    \label{fig:setup}
    \vspace{-6mm}
\end{center}
\end{figure*}

\subsubsection{Training}
We use the Proximal Policy Optimization (PPO) algorithm~\cite{schulman2017proximal} to train our meta-controller. Both the critic and actor networks include a 5-cell LSTM network and each cell's hidden size is 64. Each LSTM is followed by a fully-connected layer with 64 hidden neurons. The output dimensions of critic and actor are 1 and 2 respectively. 

Controlling these meta-parameters to strike the right balance is very convoluted and there are no straightforward intuitions, even for humans. So we use a sparse reward to avoid imposing any human prior knowledge or biasing the policy in the wrong direction. For each episode, a reward of 1 is given only if the robot arm successfully grasps and lifts the object. The episode is terminated if an unwanted collision happens, the object is knocked over, or the object has reached the end of the trajectory. We use a discount factor $\gamma = 0.95$ to incentivize the policy to obtain the reward fast, leading to short episode lengths and small dynamic grasping time. 

We formulate both meta-parameters as Gaussian Distributions and our actor network is predicting the mean of these Gaussians. The standard deviation is decreased from 1 to 0.05 linearly with the number of training steps, using a rate of 0.05 per 5000 episodes, and remains constant after reaching 0.05. The training takes around 150,000 episodes to converge (roughly 24 hours). 
The training plot with 5 random seeds is shown in Fig.~\ref{fig:trainin_plot}.
\section{Experiments}
\label{sec:exp}

\begin{table*}
    \addtolength{\tabcolsep}{-1pt}
    \footnotesize
    \centering
    \begin{tabular}{c|cc|cc|cc|cc}
    \toprule
    \multirow[b]{3}{*}{\textbf{Method}} & \multicolumn{2}{c|}{\textbf{Training}} & \multicolumn{6}{c}{\textbf{Generalization}} \\
    \cmidrule{2-9}
    & \multicolumn{2}{c|}{\textit{3-6 Random Blocks}} & \multicolumn{2}{c|}{\textit{7-9 Random Blocks}} & \multicolumn{2}{c|}{\textit{Household}} & \multicolumn{2}{c}{\textit{Cluttered Household}} \\
    & \makecell{Success \\ Rate} & \makecell{Grasping \\ Time (s)} & \makecell{Success \\ Rate} & \makecell{Grasping \\ Time (s)} & \makecell{Success \\ Rate} & \makecell{Grasping \\ Time (s)} & \makecell{Success \\ Rate} & \makecell{Grasping \\ Time (s)} \\
    \midrule
    Random & $0.498$ & $15.02 \pm 10.06$ & $0.391$ & $17.39 \pm 9.955$ & $0.498$ & $14.02 \pm 7.699$ & $0.258$ & $17.22 \pm 6.365$ \\
    Grid-search & $0.691$ & $12.64 \pm 10.54$ & $0.565$ & $15.78 \pm 11.03$ & $0.693$ & $13.36 \pm 8.259$ & $0.434$ & $16.90 \pm 6.953$ \\
    BO & $0.677$ & $13.21 \pm 10.88$ & $0.554$ & $16.43 \pm 10.22$ & $0.695$ & $13.32 \pm 8.373$ & $0.401$ & $15.81 \pm 6.672$ \\
    Online-IK & $0.367$ & $17.24 \pm 10.30$ & $0.295$ & $19.28 \pm 9.595$ & $0.401$ & $15.92 \pm 8.560$ & $0.138$ & $19.14 \pm 5.571$ \\
    Online-reachability & $0.163$ & $26.24 \pm 11.01$ & $0.112$ & $27.76 \pm 11.08$  & $0.102$ & $31.02 \pm 10.96$ & $0.045$ & $25.83 \pm 6.740$ \\
    MC (T) & $0.672$ & $12.57 \pm 10.25$ & $0.562$ & $15.66 \pm 10.62$  & $0.690$ & $12.42 \pm 7.497$ & $0.371$ & $15.94 \pm 6.882$ \\
    MC (L) & $0.711$ & $12.02 \pm 10.27$ & $0.613$ & $14.77 \pm 10.61$ & $0.695$ & $12.43 \pm 7.747$ & $0.481$ & $14.85 \pm 6.654$ \\
    MC (T + L) & \textbf{0.720} & \textbf{12.01 $\pm$ 9.975}& \textbf{0.666} & \textbf{13.85 $\pm$ 10.31}& \textbf{0.705} & \textbf{11.08 $\pm$ 6.694} & \textbf{0.554} & \textbf{14.18 $\pm$ 6.610} \\
    \bottomrule
    \end{tabular}
    \caption{\footnotesize\textbf{Performance under different experimental setups.} 
    Each number is reported over 2,000 trials. For \textit{MC (T)}, \textit{MC (L)} and \textit{MC (T + L)}, the number is averaged across the best checkpoints from 5 random seeds (10,000 trials in total). We report the success rate (higher values are better) and the average grasping time (lower values are better). The best performance is in bold.}
    \label{tab:results}
    \vspace{-6mm}
\end{table*}

Our experiments are designed to answer two questions: 
\begin{quote}
	\centering
	\it\textbf{
		(1) Does modulating meta-parameters at each iteration outperform fixed values?
		\\
		(2) Does learning-based meta-controller outperform heuristics-based meta-controller?}
\end{quote}
We show that our meta-controller, despite only being trained with 3-6 obstacles, can successfully generalize to 7-9 obstacles and to more realistic environments with unseen obstacle shapes that mimic warehouse and household scenarios. With these obstacles, the environment can be extremely cluttered, as shown in Fig.~\ref{fig:setup}.
The metrics that we are most interested in are (1) success rate - the ratio of episodes in which the robot successfully picks up and lifts the object over a height of 3cm, and (2) grasping time - the time taken for the object to be lifted up. 
Videos can be found on the \href{https://yjia.net/meta/}{project website}. 

\subsection{Experimental Setups}
\label{sec:setup}

We create a dynamic grasping environment in the PyBullet simulator. We use the UR5 robot arm and the Robotiq parallel jaw gripper. The target object is placed on a moving conveyor. 
Similar to \cite{akinola2021dynamic}, we design 4 categories of trajectories (linear, sinusoidal, rectangular, and circular) for the moving conveyor that can be flexibly randomized using a set of parameters. For each episode of dynamic grasping, we randomly select an object to grasp from 7 YCB objects~\cite{calli2017yale}. 
Details on trajectory parameterization and selected objects can be found in the appendix. 

We design 4 setups and each combines different trajectories and obstacles, as shown in Fig.~\ref{fig:setup}. We train our meta-controller only in the \textit{3-6 Random Blocks} setup; however, we evaluate the meta-controller in all 4 setups. 
The \textit{Household} and \textit{Cluttered Household} setups are designed to 
test the generalization ability of our meta-controller to out-of-domain scenes. A detailed explanation of these setups is as follows.  

\textbf{3-6 Random Blocks, 7-9 Random Blocks}. The obstacle poses are randomly sampled within a cuboidal region that incorporates both the trajectory and robot arm. These random blocks are guaranteed not to block the conveyor trajectory, by excluding a protected area with a height of 30cm and a width of 20cm surrounding the trajectory. 
The $x$, $y$, $z$ dimensions of obstacles are uniformly sampled from $[5\text{cm}, 15\text{cm}]$. The conveyor trajectory is sampled from all 4 trajectories.


\textbf{Household}. There are one top shelf and two side shelves between the robot arm and the object trajectory. The conveyor motion is sampled from linear, sinusoidal, and rectangular trajectories, without circular trajectories. The top shelf height is randomized between 40 - 60cm, and the side shelf locations are randomized so that an empty middle space of length 45 - 85cm is available. It is designed to have a reachable area in the middle of the trajectory while blocked at the start and end. 

\textbf{Cluttered Household}. The conveyor moves following a circular trajectory. There are 5 cylinder obstacles surrounding the trajectory and a top circular shelf obstacle covering the trajectory. The top circular shelf consists of 15 identical convex trapezoidal parts. Positions of these cylinder obstacles are randomly sampled. This is a harder and more cluttered setup compared to \textit{Household}. It is motivated by operating in extremely cluttered household and warehouse environments.

\subsection{Baselines}
We compare our approach to a variety of baselines that can be classified into two groups. (1) Assigning a fixed value to the meta-parameter across all episodes. In such a case, we can treat the meta-parameters as hyperparameters, and then it opens up the rich literature of hyperparameter tuning. We choose \textit{Grid-search} and the state-of-the-art Bayesian optimization~\cite{snoek2012practical} method. (2) Dynamically changing the meta-parameter within a single episode according to some heuristics, including \textit{Random}, \textit{Online-IK}, and \textit{Online-Reachability}. 



\textbf{Random}. This method dynamically samples a look-ahead time and a time budget uniformly from their available ranges at each iteration within a dynamic grasping episode. 

\textbf{Grid-search}. This method finds the best combination of look-ahead time and time budget through grid search. It uses the same fixed values across all episodes. It chooses 8 values evenly spaced in the action space for both the look-ahead time and time budget. We evaluate all 64 pairs with 2,000 trials per pair in \textit{3-6 Random Blocks}. The pair with the highest performance is chosen as the fixed meta-parameters. We find (2s, 1s) gives the best performance.

\textbf{BO}. This method finds the best combination of look-ahead time and time budget through Bayesian optimization. It uses the same fixed values across all episodes. We use the SMAC3 framework~\cite{JMLR:v23:21-0888} for this method, a highly optimized and flexible framework for hyper-parameter tuning. We use the random forest~\cite{breiman2001random} as its surrogate model and expected improvement~\cite{Jones1998EI} as its acquisition function. 
At each iteration, the chosen combination is evaluated in \textit{3-6 Random Blocks} for 1,000 trials.
We optimize for 300 iterations and find (0.82s, 0.78s) gives the best performance. 

\textbf{Online-IK}. This method discretizes the predictable range $[0s, 8s]$ into $N$ segments. 
It computes the predicted poses at the centers of these segments and selects the look-ahead time of the pose that is farthest away with a valid IK solution. Choosing a pose that is far reserves enough time for motion and grasp planning. 
In general, choosing a predicted pose that far incurs higher error in prediction accuracy. However, for this baseline in practice, due to the large computation time, reserving more time plays a more important factor. 
The number of segments $N$ brings another trade-off into the dynamic grasping system. Higher $N$ allows for a more accurate IK estimate of the entire predictable range, but it also takes a longer time. We found through experiments that $N=3$ strikes the right balance for this trade-off. For time budget, we use the value found in \textit{Grid-search}.

\textbf{Online-reachability}. \label{sec:reachability} 
This method is an extension from \textit{Online-IK}. After computing the IK for the $N$ predicted poses, for those poses with a valid IK solution, this method further computes their reachability values, namely, how easy it is to plan a collision-free path to those poses. 
Given a time-budget $T_T$, to compute the reachability of a particular predicted pose, it attempts to plan a motion towards that pose $M$ times. $t_i$ denotes the motion planning time for the $i$-th attempt. Then the reachability value $R$ is computed as $$R = 1 - \frac{\sum_{i=1}^{M} t_i}{M T_T},\,\, 0 < t_i \leq T_T$$ 
When all the attempts fail, namely, the entire time budget is used up, then the reachability value of that pose is 0. 
This method chooses the look-ahead time such that the predicted pose has the largest reachability. It uses the fixed time budget found in \textit{Grid-search}. 


\textbf{MC (T), MC (L)}. These are ablations that learn to control only time budget \textit{or} look-ahead time, respectively, while the other one is the fixed value found in the \textit{Grid-search}.

\textbf{MC (T + L)}. This is our proposed method. 

\begin{figure*}
    \centering
    \includegraphics[width=\linewidth]{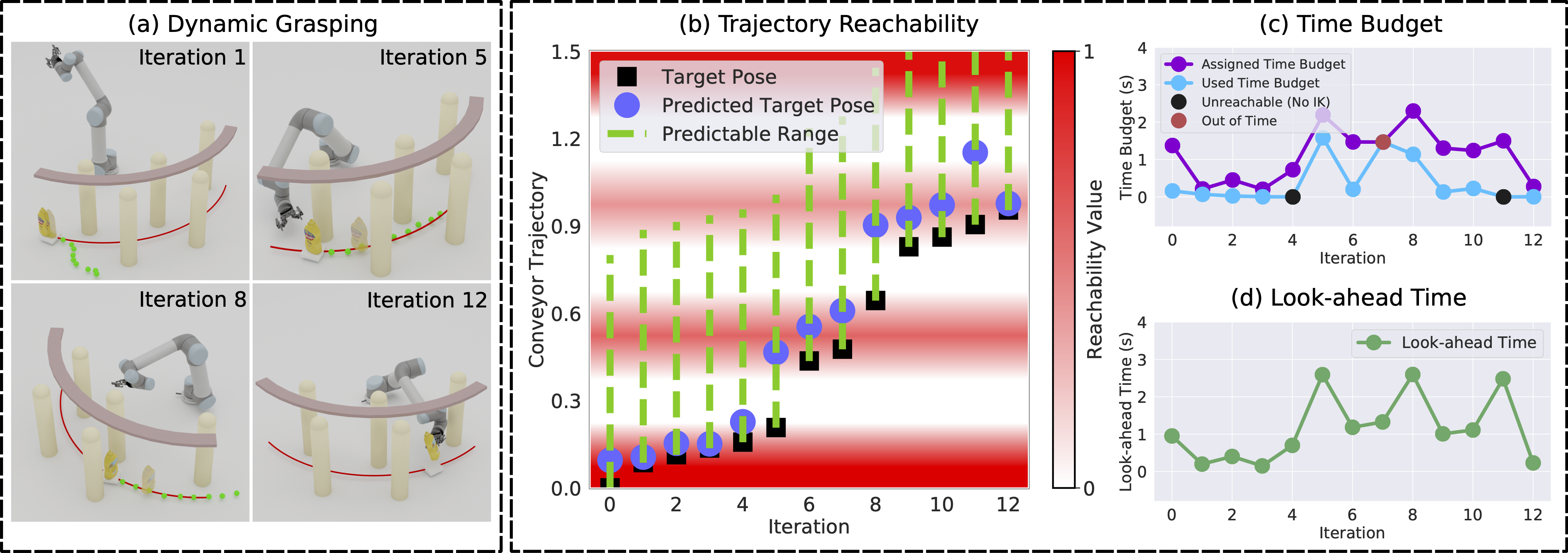}
    \vspace{-5mm}
    \caption{\footnotesize\textbf{Meta-controller demonstration on the \textit{Cluttered Household} setup.} (a) Scenes at different iterations within the same episode. 
    Green dots indicate the predicted trajectory for the whole predictable range. 
    The semi-transparent object indicates the predicted pose at the look-ahead time. 
    (b) Trajectory reachability visualization. The whole trajectory and its offline-computed reachability 
    are projected vertically onto the $y$-axis of the plot. 
    (c) The assigned time budget by our meta-controller for each iteration. (d) The assigned look-ahead time by our meta-controller for each iteration.}
    \label{fig:qualitative}
    \vspace{-3mm}
\end{figure*}

\begin{figure}[h!]
    \centering
    \includegraphics[width=\linewidth]{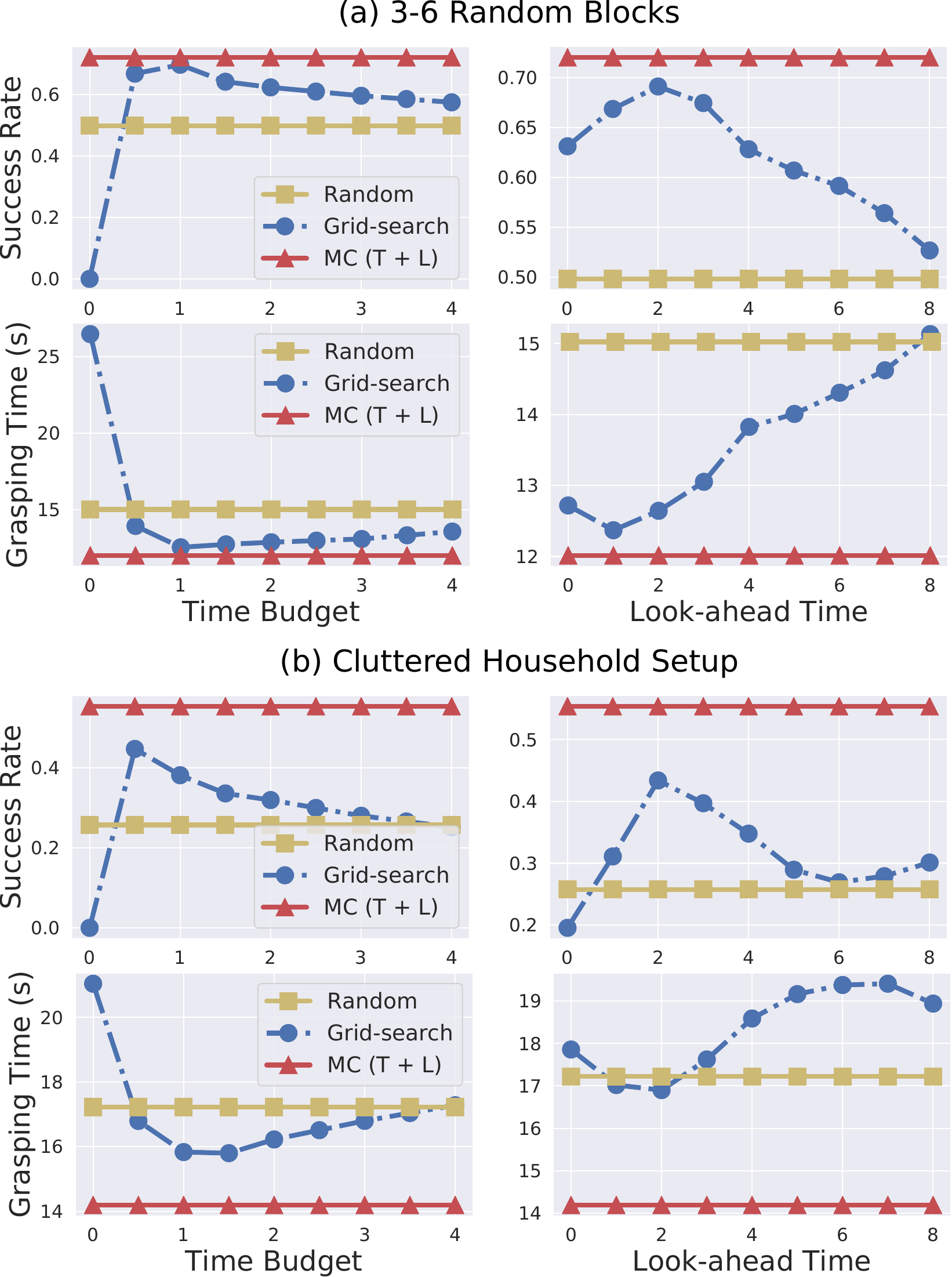}
    \vspace{-5mm}
    \caption{\footnotesize\textbf{Comparison with different fixed meta-parameters in \textit{Grid-search}.} (a) \textit{3-6 Random Blocks}. (b) \textit{Cluttered Household}. For each setup: the top row is the success rate, and the higher is better. The bottom row is the grasping time, and the lower is better. Each number is averaged over 2,000 trials. Our meta-controller outperforms both \textit{Random} and \textit{Grid-search}, and the gap becomes more significant as the environment becomes more cluttered and challenging.}
    \label{fig:compare}
    \vspace{-5mm}
\end{figure}

\subsection{Performance Analysis and Discussion}

As shown in Table~\ref{tab:results}, our meta-controller achieves a higher success rate and lower dynamic grasping time, due to three factors. (1) It can reason about the reachable workspace and through dynamically controlling the look-ahead time and time budget, it maintains the predicted pose and the planned motion within the most reachable region. (2) It learns to generate a small look-ahead time when the predicted trajectory is not accurate. (3) It learns to produce a small but sufficient time budget for motion planning. 

We demonstrate this behavior using an example from \textit{Cluttered Household}, as shown in Fig.~\ref{fig:qualitative}. At the beginning of each episode, our meta-controller tends to output a small look-ahead time as the predictor does not have enough past information to generate correct predicted poses. This behavior can be seen in iteration 1 of Fig.~\ref{fig:qualitative}(a). 
Due to these cylinder obstacles, there are 4 small reachable regions, interwoven with unreachable regions. Our meta-controller is able to maintain the predicted pose within these reachable regions most of the time. When the object moves close to the edge of a reachable region, our meta-controller produces a far predicted pose that falls right into the next reachable region. As the target object is moving within the reachable region, our meta-controller gradually reduces the look-ahead time to maintain the predicted pose within that region. As for time budget, as shown in Fig.~\ref{fig:qualitative}(c), the meta-controller produces a time budget that approximates the amount of time a successful path can be planned. 


On the contrary, \textit{Grid-search} fails to react to scene information, despite being the strongest baseline without a meta-controller. Using the same example in Fig.~\ref{fig:qualitative}, neither large nor small fixed look-ahead time will maintain the predicted pose within the reachable region for long enough. As a result, the robot fails constantly to move closer to the target. 
We further compare our method to \textit{Grid-search} on different fixed look-ahead times and time budgets in the \textit{3-6 Random Blocks} and \textit{Cluttered Household}, as shown in Fig.~\ref{fig:compare}. We keep one of the meta-parameters in the optimal pair and then vary the other. These results show how different fixed meta-parameters can significantly impact performance. However, our meta-controller outperforms all fixed values by modulating the meta-parameters at different iterations, based on the scene information. The gap is further exacerbated when the environment becomes more cluttered as we go from \textit{3-6 Random Blocks} to \textit{Cluttered Household}.

\textit{BO} struggles to construct an accurate surrogate model due to the randomness of the environments. The success of one dynamic grasping episode can depend on many factors, such as the sampled object, obstacle, and trajectory. Over 300 optimization iterations and 1,000 trials per iteration are required to converge properly, which is already two times the number of episodes required by our meta-controller. 

\begin{figure}
    \centering
    \includegraphics[width=\linewidth]{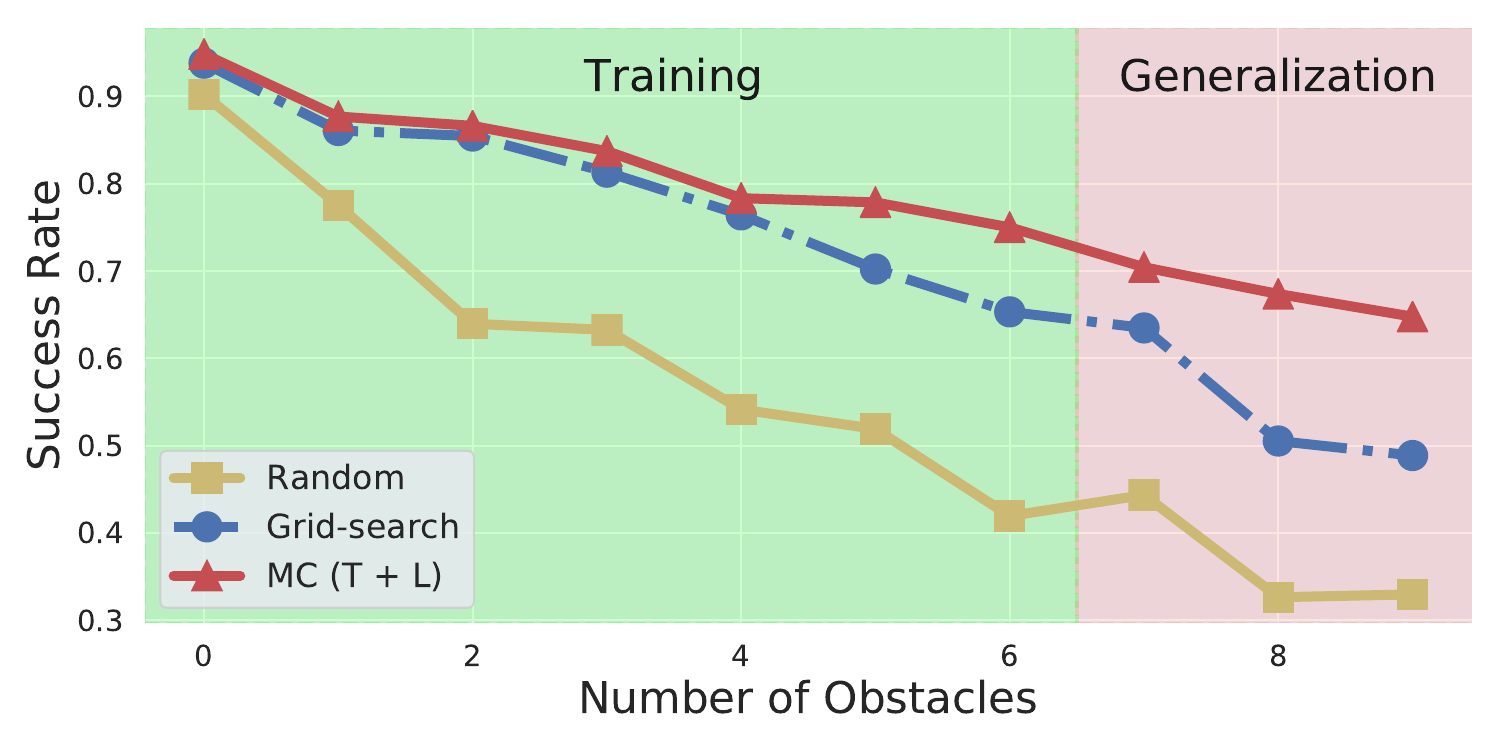}
    \vspace{-5mm}
    \caption{\footnotesize\textbf{Success rate with the number of obstacles.} We fix the number of obstacles and evaluate each method for 2,000 trials. 
    The meta-controller is only trained on \textit{3-6 Random Blocks}.
    A larger gap in success rate is observed as we go beyond the number of obstacles during training.}
    \label{fig:num_obstacles}
    \vspace{-6mm}
\end{figure}

\textit{Online-IK} and \textit{Online-reachability} attempt to dynamically assign meta-parameters by explicitly utilizing IK information but their performance is largely affected by the online computation delay. In the dynamic grasping experiments, the target object most likely has moved away as the online computation finishes. \textit{Online-reachability} has an even larger delay because it also considers reachability. Thus, it performs even worse than \textit{Online-IK}. Our learning-base meta-controller also infers the reachability space, but implicitly and runs much faster at inference time. As for the ablations, \textit{MC (L)} constantly outperforms \textit{MC (T)}, indicating that look-ahead time plays a more important role in dynamic grasping.


The advantage of our meta-controller becomes more pronounced with more cluttered environments. Our method is 18\% and 28\% higher on success rate than the fixed meta-parameters found through \textit{Grid-search} in the \textit{7-9 Random Blocks} and \textit{Cluttered Household} respectively. In cluttered environments, the feasible workspace of the robot becomes extremely limited, and the ability to reason about the reachability space and maintain the arm in those areas becomes critical. We plot the success rate of \textit{Random}, \textit{Grid-search} and \textit{MC (T + L)} as a function of the number of obstacles in Fig.~\ref{fig:num_obstacles}. We can see that without any obstacles, all three methods achieve above 90\% success rate. As the number of obstacles exceeds the training number, the gap between ours and the other two baselines becomes more significant. 
\section{Conclusion}
We introduce a dynamic grasping pipeline with a learned meta-controller that dynamically assigns the look-ahead time and time budget. We learn the meta-controller through reinforcement learning with a sparse reward and evaluate it on various experimental setups. The results show that the learned meta-controller can drastically improve the dynamic grasping performance.
We believe our method can be applied to many other robotic dynamic applications consisting of individual submodules and meta-parameters.

\bibliographystyle{IEEEtran}
\bibliography{references}

\clearpage
\newpage
\section{Appendix}
\subsection{Object Pose Predictor Architecture}
\begin{figure}[h]
    \centering
    \includegraphics[width=\linewidth]{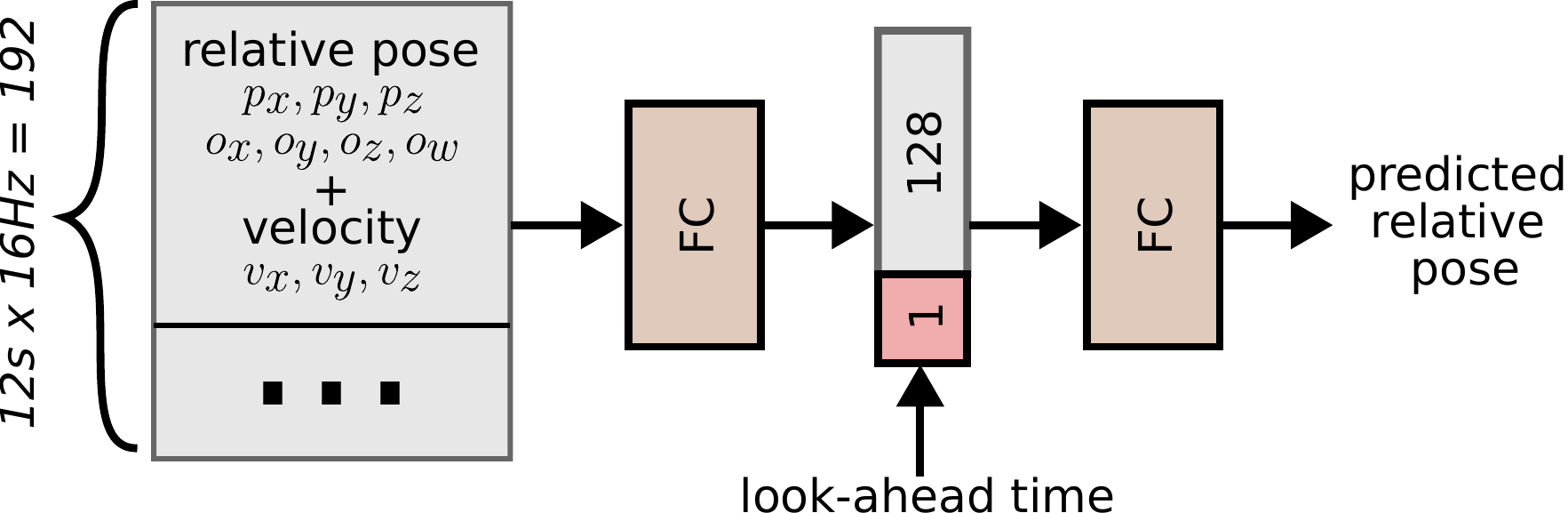}
    \vspace{-5mm}
    \caption{\footnotesize\textbf{Object pose predictor architecture.} The sequence of past poses and velocities is sampled from the past 12 seconds at 16Hz. The first FC (fully-connected) network extracts a 128-dimensional feature, and the second FC network generates the predicted pose from the extracted feature and look-ahead time. All poses are relative to the previous pose.}
    \label{fig:pose_predictor}
    \vspace{-5mm}
\end{figure}

The architecture of our pose predictor is shown in Fig.~\ref{fig:pose_predictor}. The pose predictor first extracts a 128-dimensional feature from the history buffer, and then the look-ahead time is appended to this feature, passed together through another fully-connected network to get the final predicted pose. We choose not to use recurrent architectures such as long short-term memory (LSTM) to achieve faster inference time, such that we can compute the whole predicted trajectory for the predictable range in parallel. We also tried a Transformer~\cite{vaswani2017attention} architecture, but it does not outperform the current version based on fully-connected networks. We reach the same conclusion as stated in~\cite{zeng2022transformers}, which highlights that the Transformer architecture may not outperform in time-series forecasting tasks.

\subsection{Trajectories and Objects}
\begin{figure}[h]
    \centering
    \includegraphics[width=\linewidth]{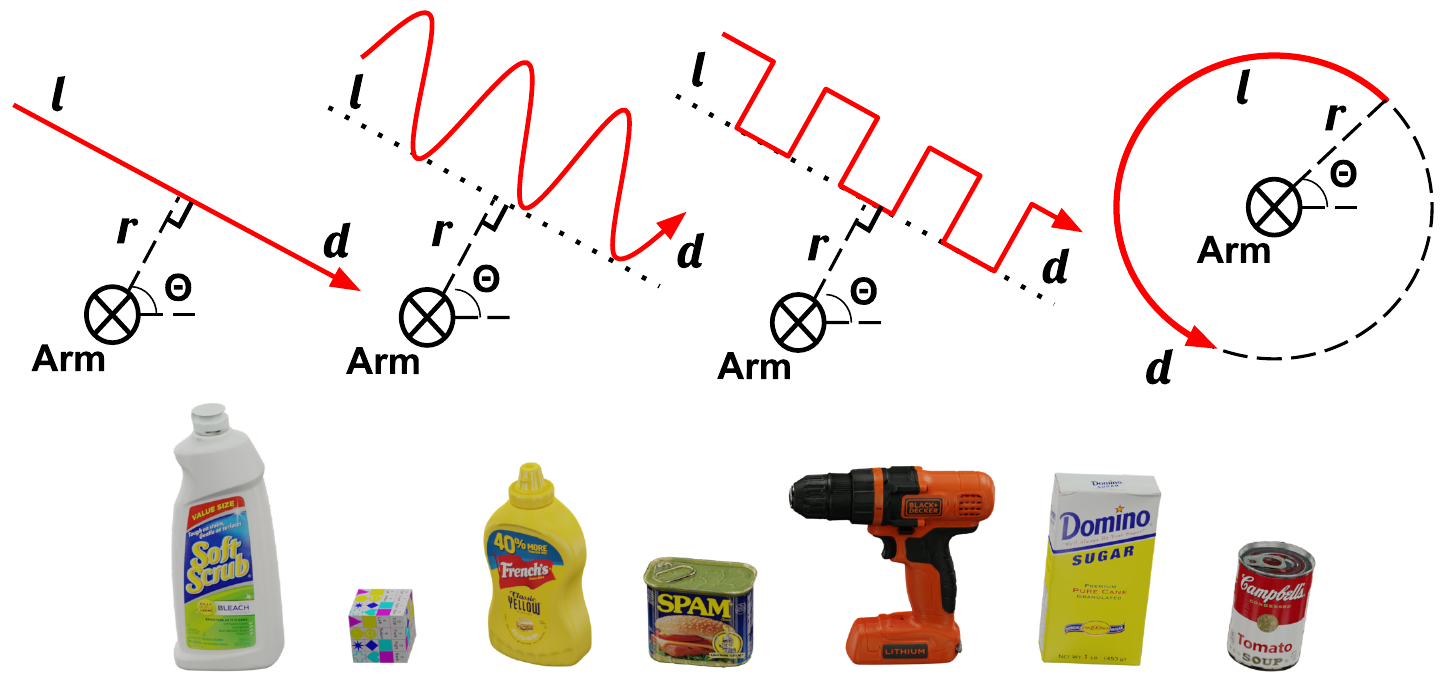}
    \vspace{-5mm}
    \caption{\footnotesize\textbf{Randomized trajectories and selected graspable objects.} Top Row: A bird's eye view of linear, sinusoidal, rectangular, and circular conveyor belt motion generation process. Each random motion is parameterized by angle $\theta$, distance $r$, length $l$, and direction $d$. The cross indicates the position of the robot base. The red line shows the motion of the conveyor belt, with an arrow indicating the direction. The horizontal dashed line at the robot base indicates the positive $x$-axis of the world frame. Bottom row: 7 objects from the YCB dataset are selected as the graspable target objects for our experiments.}
    \label{fig:trajectories}
    \vspace{-6mm}
\end{figure}

How the 4 types of trajectories are parameterized and the set of selected objects are shown in Fig.~\ref{fig:trajectories}. $\theta$ specifies the counter-clockwise angle of the trajectory, $r$ is the distance from the trajectory to the robot arm base, $l$ is the length of the trajectory (for rectangular and sinusoidal trajectories, $l$ is the straight distance from start to end), and $d \in \{+1, -1\}$ indicates the direction of the motion, where $+1$ means counter-clockwise and $-1$ means clockwise. The speed $v$ of the motion is randomly sampled but remains constant through an episode. 

We select this set of objects to include both concave and convex shapes and also to ensure that the size of the object fits in the gripper. The sampling range of each trajectory parameter is shown in Table~\ref{tab:trajectories}. We design these ranges such that at least part of the trajectory is reachable for the robot arm, assuming no obstacles.

\begin{table}[h]
    \setlength\tabcolsep{4.5pt}
    \centering
    \begin{tabular}{c|ccccc}
    \toprule
    Trajectory & $\theta$ & $r$ (m) & $l$ (m) & $d$ & $v$ (cm/s) \\
    \midrule
    Linear & [0\degree, 360\degree] & [0.35, 0.65] & 1 & $\{+1, -1\}$ & [2, 6] \\
    Sinusoidal & [0\degree, 360\degree] & [0.35, 0.65] & 1 & $\{+1, -1\}$ & [2, 6]\\
    Rectangular & [0\degree, 360\degree] & [0.35, 0.65] & 1 & $\{+1, -1\}$ & [2, 6]\\
    Circular & [0\degree, 360\degree] & [0.65, 0.9] & 1.5 & $\{+1, -1\}$ & [5, 10] \\
    \bottomrule
    \end{tabular}
    \caption{\textbf{Parameter sampling range for each conveyor trajectory.}}
    \label{tab:trajectories}
    \vspace{-5mm}
\end{table}

\subsection{Justification on Sim2Real Transfer}
We are not able to provide real-world experiments. However, we think that the sim2real gap of this work is minimal, and the performance gain from the meta-controller will be consistent regardless of the sim2real gap. There are several reasons for it. (1) The largest gap when transferring to the real robot is the vision system that estimates the object poses and bounding box dimensions. Recent perception systems such as DOPE~\cite{tremblay2018deep} and Gen6D~\cite{liu2022gen6d} can obtain such information fast and reliably. We also add significant Gaussian noises to the poses and bounding box dimensions to account for such perception errors. Given the same magnitude, pure random noise is more challenging than noise from realistic perception systems with fixed observation-to-noise mapping. (2) Given the estimated poses and bounding box dimensions, there is no sim2real gap in other components of our pipeline, such as object pose predictor, grasp planner, motion planner, and the meta-controller. (3) Our simulated environment is under realistic timing. Time spent in all submodules is compensated with the target object motion and we retime the planned arm motion to make sure it matches the real-robot speed.

\end{document}